\documentclass[letterpaper, 10 pt, conference]{ieeeconf}  

\IEEEoverridecommandlockouts                              

\usepackage{graphicx}
\usepackage{amsmath}
\usepackage{amsfonts}
\usepackage{bm}
\usepackage{booktabs}
\usepackage{color}
\usepackage{url}
\usepackage{amsmath}

\newcommand{\figref}{Figure~\ref}
\newcommand{\tabref}{Table~\ref}
\newcommand{\etal}{\textit{et al}.}
\newcommand{\ie}{\textit{i}.\textit{e}.}

\title{\LARGE \bf
APriCoT: Action Primitives based on Contact-state Transition for In-Hand Tool Manipulation
}

\author{Daichi Saito$^{1,2}$, Atsushi Kanehira$^{2}$, Kazuhiro Sasabuchi$^{2}$, Naoki Wake$^{2}$,\\ Jun Takamatsu$^{2}$, Hideki Koike$^{1}$ and Katsushi Ikeuchi$^{2}$
\thanks{$^{1}$Authors are with Department of Computer Science,
        Tokyo Institute of Technology, Tokyo, Japan, {\tt\small saito.d.ah@m.titech.ac.jp}}%
\thanks{$^{2}$Authors are with Applied Robotics Research, Microsoft, Redmond, WA, USA}%
}

\begin{document}

\maketitle
\thispagestyle{empty}
\pagestyle{empty}

\begin{abstract}
In-hand tool manipulation is an operation that not only manipulates a tool within the hand (\ie, in-hand manipulation) but also achieves a grasp suitable for a task after the manipulation. 
This study aims to achieve an in-hand tool manipulation skill through deep reinforcement learning.
The difficulty of learning the skill arises because this manipulation requires (A) exploring long-term contact-state changes to achieve the desired grasp and (B) highly-varied motions depending on the contact-state transition. (A) leads to a sparsity of a reward on a successful grasp, and (B) requires an RL agent to explore widely within the state-action space to learn highly-varied actions, leading to sample inefficiency.
To address these issues, this study proposes Action Primitives based on Contact-state Transition (APriCoT). 
APriCoT decomposes the manipulation into short-term action primitives by describing the operation as a contact-state transition based on three action representations (\textit{detach}, \textit{crossover}, \textit{attach}). 
In each action primitive, fingers are required to perform short-term and similar actions.
By training a policy for each primitive, we can mitigate the issues from (A) and (B).
This study focuses on a fundamental operation as an example of in-hand tool manipulation: rotating an elongated object grasped with a precision grasp by half a turn to achieve the initial grasp. 
Experimental results demonstrated that ours succeeded in both the rotation and the achievement of the desired grasp, unlike existing studies. 
Additionally, it was found that the policy was robust to changes in object shape. 
\end{abstract}

\section{Introduction}
\begin{figure}[t]
  \centering
  \includegraphics[width=\columnwidth]{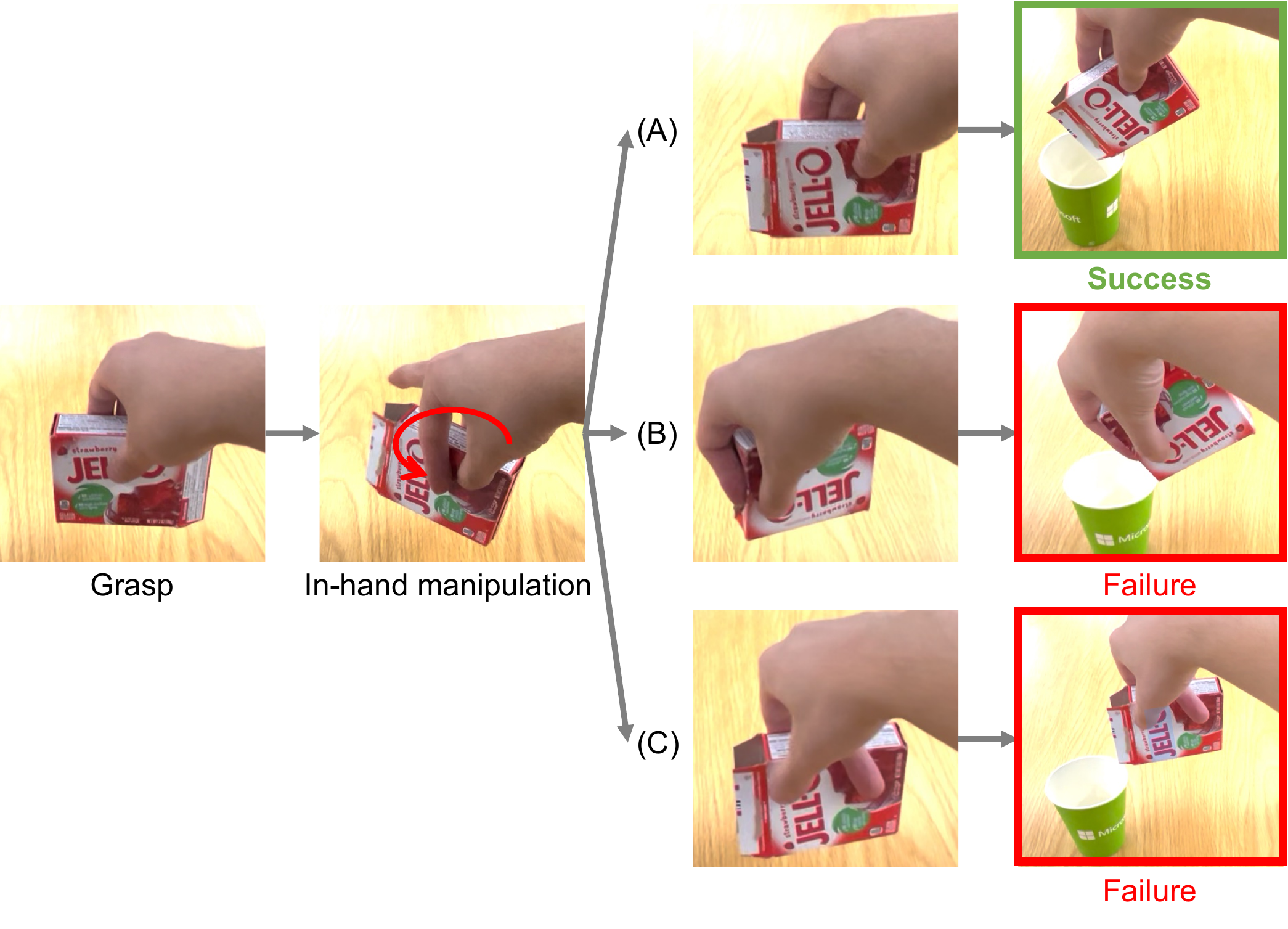}
  \caption{An example of a task after in-hand tool manipulation: pouring the contents of a box into a cup by shaking it with fingers. 
  To orient an opening of the box towards the cup, the hand must rotate the box counterclockwise (red arrow in the second image from the left).
  To successfully complete this task, the hand should grasp the object as shown in (A) after manipulation. If the object is grasped as in (B) or (C), issues arise such as the fingers blocking the opening or the inability to properly change the object orientation.}
  \label{fig:examples}
  \vspace{-6mm}
\end{figure}

In-hand tool manipulation~\cite{zarrin2023hybrid, gordon2023online} is an operation that not only manipulates a tool within the hand (\ie, in-hand manipulation) but also achieves a grasp-type~\cite{feix2015grasp} suitable for a task after the manipulation.
Since the inappropriate grasp may cause the tool not to function as expected and task failure, realizing a suitable grasp through the manipulation is crucial for the subsequent task. For instance, when shaking a box with fingers to pour its contents into a cup, we should ensure an opening of the box faces the cup and the box is held with a precision grasp (\figref{fig:examples}-(A)). If the object is not properly grasped, the opening is blocked by fingers (\figref{fig:examples}-(B)), or it is impossible to change the object orientation (\figref{fig:examples}-(C)). 
The in-hand tool manipulation skill is essential for a dexterous operation with various types of tools in daily life. 
The acquisition of the ability on a single hardware is necessary for realizing a general-purpose robot.

The objective of this study is to achieve the manipulation skill through deep reinforcement learning (DRL). DRL allows for the acquisition of policies robust to various uncertainties, such as object shape variations and observed joint angle errors, making it the de facto standard in the field of in-hand manipulation~\cite{andrychowicz2020learning, handa2023dextreme, yin2023rotating, chen2022system, qi2023hand, qi2023general}. 

\begin{figure}[t]
  \centering
  \includegraphics[width=\columnwidth]{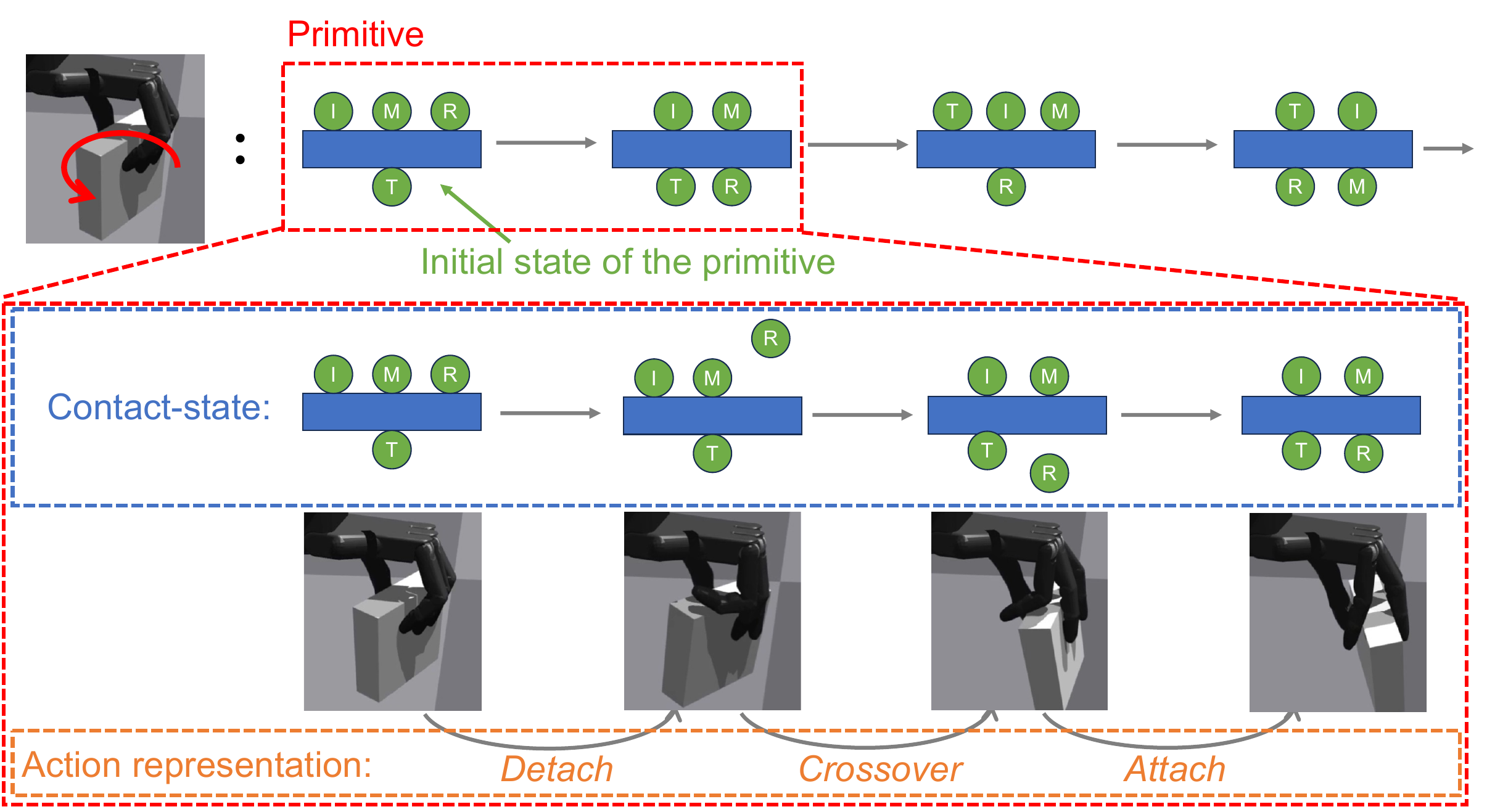}
  \caption{The explanation of action representations and primitives. 
  I, M, R, and T represent the index finger, middle finger, ring finger, and thumb, respectively.
  This figure shows an example of a contact-state transition in case of rotating the box counterclockwise.
  \textit{Detach}, \textit{crossover}, and \textit{attach} are the action representations to transition the contact-state. The initial contact-state of a primitive is set to the most stable one where all fingers are in contact with the object.
  }
  \label{fig:primitives}
  \vspace{-6mm}
\end{figure}

Although DRL has the potential to facilitate in-hand tool manipulation, learning this skill remains difficult.
The difficulty arises from the need for (A) long-term contact-state changes and 
(B) highly-varied motions depending on the contact-state transition
for the manipulation. With regard to (A), fingertip positions relative to the object must change many times to achieve the desired grasp, involving long-term contact-state transitions. As for (B), required motions vary depending on current fingertip positions. For example, in the left situation of \figref{fig:examples}, if the ring finger moves to the opposite side of the object, the others must adjust the object orientation as shown in the lower of \figref{fig:primitives}. 
Conversely, when one of the others moves to the opposite, actions of this finger significantly differ from those in the lower of \figref{fig:primitives}.

When learning an in-hand tool manipulation skill using DRL, two main issues corresponding to (A) and (B) arise. The first issue, due to (A), is a sparsity of a reward for a successful grasp over time. 
This sparsity makes it hard to map the reward to previous actions predicted by a policy and thus the learning becomes difficult.
The second issue, stemming from (B), is that an RL agent must widely explore within the state-action space to learn the highly-varied actions. 
As a result, the number of training samples to find the optimal policy significantly increases.
A possible solution for the sparsity is adding a dense reward such as one on a discrepancy between the joint angles of the current hand configuration and those of the target grasp~\cite{qi2023hand, qi2023general}. However, this reward limits the exploration within the state-action space and further worsens sample efficiency. 
Thus, learning manipulation characterized by both the reward sparsity and sample inefficiency presents significant challenges.

These issues can generally be solved by decomposing a long-term motion into short-term motions~\cite{gudimella2017deep, clegg2018learning, li2020learning, chen2023sequential}. Although in-hand tool manipulation appears to involve long-term and complex motions overall, it consists of a combination of three abstract action representations to describe contact-state changes: \textit{detach}, \textit{crossover}, and \textit{attach}~\cite{vinayavekhin2013representation}. \textit{Detach} is to move a finger away from the object. \textit{Crossover} is a motion where a finger changes from one side of the object to the other side of the object. \textit{Attach} is to bring a finger into contact with the object (\figref{fig:primitives}).
The desired grasp can be defined as the contact-state~\cite{kang1993toward}, and it is reasonable to regard in-hand tool manipulation, which is the operation to change the grasp, as the contact-state transition.
Thus, we can describe the entire manipulation by combining the representations, and decompose the manipulation into short-term motions based on this description. Such temporal decomposition can alleviate the issue due to the characteristic (A).
In addition, we can mitigate the problem from the characteristic (B) by spatially decomposing so that the fingers behave similarly within each motion and separately training the policy for each motion.

This study proposes the learning of in-hand tool manipulation using Action Primitives based on Contact-state Transition (APriCoT). By sequentially executing the action primitives, we achieve the manipulation. That is, we consider the transitions to achieve the desired grasp, decompose the manipulation into short-term primitives based on the transition, and introduce the learning of each action primitive.
This study defines the contact-state based on not only the arrangement of fingers but also grasp stability~\cite{ferrari1992planning} and manipulability.
In addition, the initial and final contact-states of each primitive are set to the most stable one where all fingers are in contact with the object.
In this case, the primitives make transitions between these contact-states by the series of detach-crossover-attach.
Since the most stable contact-state is less affected by dynamics such as object velocity and the friction forces between the object and the fingers, this contact-state is easily realized. Thus, this state can be stably sampled in simulations, which improves the efficiency of initial-state sampling.
Additionally, training the policy to learn a set of representations rather than each individually enhances the ability of the policy to perform a smooth motion, thereby promoting stable and successful manipulation.
The policies are trained using a step-wise reward, promoting execution in the detach-crossover-attach order. This reward can be used for all primitive policies. 

We target an example of in-hand tool manipulation where the hand rotates an elongated object half a turn to achieve a precision grasp, as shown in \figref{fig:examples}-(A). 
This operation has long-horizon contact-state transition, and is frequently utilized for daily-life tools including wrenches and cups with lids.
Simulation experiments using a multi-fingered hand demonstrated that existing methods focusing solely on in-hand manipulation or targeting pinchable objects fail to achieve the desired grasp or fail the rotation itself. In contrast, ours achieves the desired grasp after the rotation with a $90\%$ success rate for objects of various shapes. 

The contributions of this study are as follows:
\begin{itemize}
    \item Proposed APriCoT for learning the in-hand tool manipulation skill.
    \item Introduced reward design focusing on \textit{detach}, \textit{crossover}, and \textit{attach}, which is reused for training all primitive policies.
    \item Demonstrated that our method is effective for the manipulation with long-term contact-state changes and highly-varied actions.
\end{itemize}
Although this study targets primitives for four contact-state transitions, our method can be easily applicable to various operations by increasing the number of reusable primitives. 
For instance, transitioning from a precision grasp to a power grasp~\cite{feix2015grasp} after rotating an object half a turn and vice versa becomes feasible. Even with more primitives, the reward design remains reusable, avoiding significant increases in reward-design costs. Thus, our method is potentially applicable to other operations. This study is a first step toward realizing diverse in-hand tool manipulation through the combination of primitives.

\section{Related Work}
In-hand manipulation has been extensively researched over the past few decades~\cite{okamura2000overview}. Early studies relied on precise models of a robot hand and an object for motion planning~\cite{han1998dextrous, rus1999hand, mordatch2012contact, bai2014dexterous}. However, uncertainties of the models in real limits their practical applicability.
To address these challenges, this study utilizes DRL to train the policy to learn object manipulation robust to the uncertainties.

Advancements in DRL~\cite{ibarz2021train} have accelerated the policy training across a wide range of domains, including robot locomotion, grasping, and object manipulation~\cite{lee2020learning, Rajeswaran-RSS-18, wu2023daydreamer}. The field of in-hand manipulation has also adopted DRL~\cite{andrychowicz2020learning, handa2023dextreme, yin2023rotating, chen2022system, qi2023hand, qi2023general}.
In-hand tool manipulation requires not only in-hand manipulation but also the achievement of the desired grasp after the manipulation. Previous research has focused on acquiring the former skill, such as rotating various shaped objects~\cite{andrychowicz2020learning, handa2023dextreme, yin2023rotating, chen2022system, melnik2021using, qi2023hand, qi2023general, yang2024anyrotate, pitz2023dextrous, khandate2022feasibility}. These studies have primarily designed a reward only concentrating on in-hand rotation, and not ensured realizing the desired grasp.
Although there are several studies that focus on learning in-hand tool manipulation~\cite{zarrin2023hybrid, gordon2023online}, they require precise models of a hand and object, making real-world application difficult. The research by Qi \etal~\cite{qi2023hand, qi2023general} potentially has the ability to maintain a precision grasp after rotating pinchable objects, however, their method is limited to operations with a narrow state-action space, failing to achieve an operation that requires exploring a broader space.
This study addresses such limitations by decomposing the manipulation into several primitives and by separately training each primitive policy. This approach simplifies the acquisition of the manipulation.

Existing researches enable the learning of long-horizon tasks by decomposing them into shorter motions~\cite{konidaris2009skill, gudimella2017deep, clegg2018learning, li2020learning, chen2023sequential, dalal2021accelerating}. For example, some studies have segmented a sequence of tasks from grasp to manipulation, and designed policies or controllers for each motion~\cite{gudimella2017deep, chen2023sequential, dalal2021accelerating}. Other research has divided in-hand manipulation in a 2-dimensional space into multiple actions, and designed controllers for each action assuming the known model~\cite{li2020learning}.
This study focuses on in-hand tool manipulation in a 3D space. We consider short-term action primitives for the manipulation and design the learning for these primitives. 

\section{Method}
This study aims to achieve in-hand tool manipulation skills using DRL. The primary challenges in learning this kind of manipulation are:
\begin{itemize}
\item The necessity for long-term exploration to achieve the desired grasp.
\item The highly-varied actions depending on the contact-state transition, necessitating the exploration of a wide state-action space.
\end{itemize}
To address these challenges, we propose Action Primitives based on Contact-state Transition (APriCoT). 
The operation is decomposed into several primitives so that the fingers are required to perform short-term and similar motions within each primitive. This temporal and spatial decomposition mitigates these issues.
We achieve the manipulation by sequentially executing primitive policies.

\subsection{Contact-state transition}
\begin{figure}[t]
  \centering
  \includegraphics[width=\columnwidth]{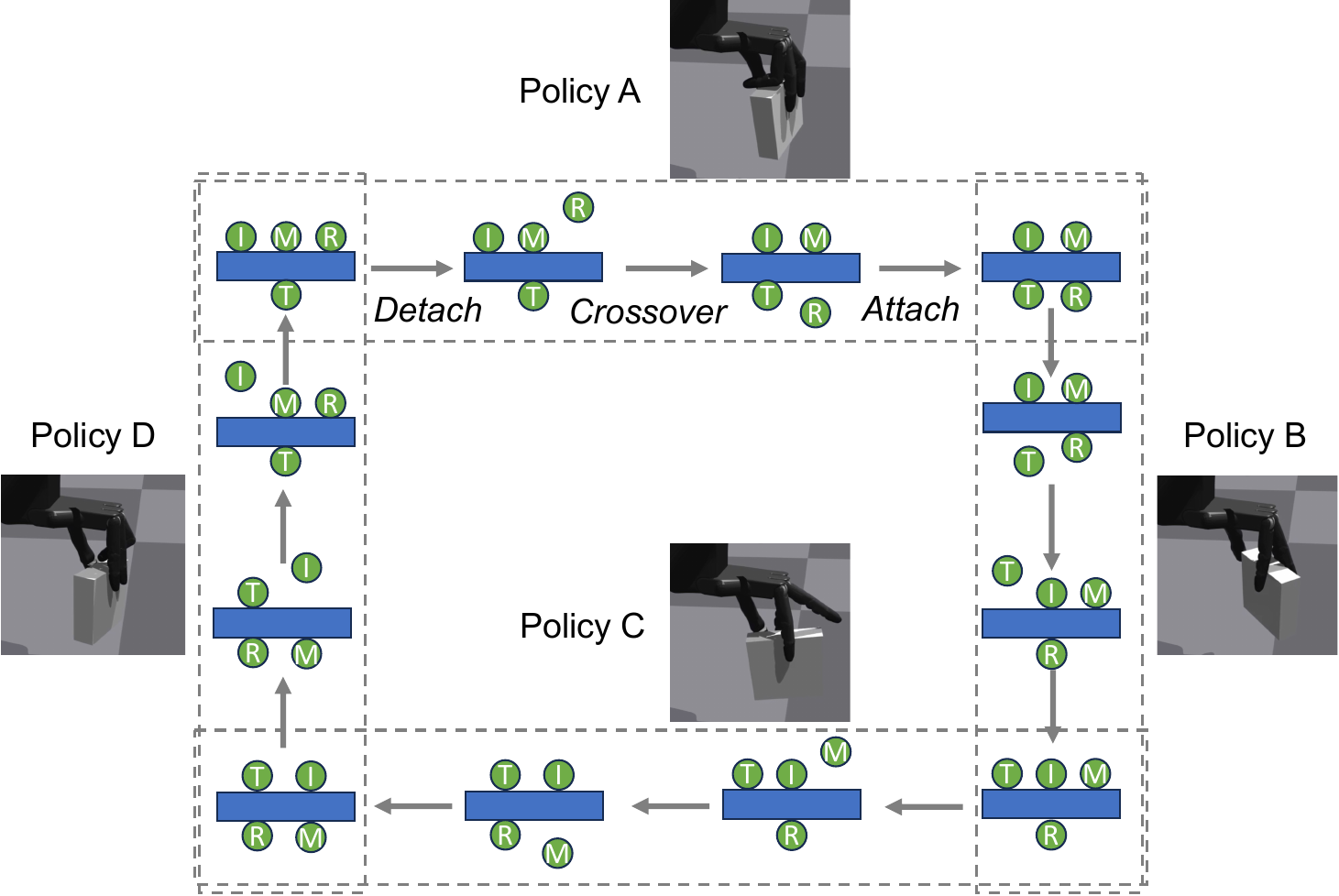}
  \caption{The contact-state transitions in the targeted manipulation. Each section enclosed by dashed lines represents an action primitive.}
  \label{fig:transition}
  \vspace{-6mm}
\end{figure}

To decompose in-hand tool manipulation, we design the contact-state transition-graph of the manipulation. 
This graph has contact-states as nodes and contact-state transitions as edges.
We construct a transition-graph based on three action representations: \textit{detach}, \textit{crossover}, and \textit{attach}.
Before the construction, we design the contact-state in the graph. This study defines the contact-state based on not only the finger arrangement but also the grasp stability and manipulability. The stability is evaluated with the closure~\cite{ferrari1992planning}. The manipulability ensures that the fingers do not limit each other's range of motion after the transition. Specifically, regarding the grasp stability, we require that a prehensile grasp is achieved and that at least three fingers are in contact with the object. As for the manipulability, the fingers must be positioned clockwise in the order of thumb, index, middle, and ring finger to avoid entanglement of the fingers.
The graph is then constructed by connecting the contact-states that can transition through detach, crossover, and attach motions. 
Under these conditions, the transition of contact-states for the targeted manipulation is shown in \figref{fig:transition}. Initially, the ring finger detaches from the object, performs crossover, and attaches to change the contact-state. Next, the same motions are performed sequentially by the thumb, middle, and index fingers to achieve the desired grasp. 

We set the initial and final contact-states of each primitive as the most stable one where all fingers contact the object. 
This stability, being less affected by dynamics such as object velocity and friction forces, makes it easier to achieve such contact-state. Thus, the contact-state can be stably sampled in simulations, which improves the efficiency of initial-state sampling.
By training a policy to learn a set of representations rather than each individually, we can promote the ability of the policy to perform a smooth motion.
The smoothness leads to stable and successful manipulation.
For instance, learning only the detach may cause the finger to move excessively away from the object, adding unnecessary motion before the crossover. Whereas, training the policy to execute the detach-crossover-attach sequence enables actions based on subsequent representation, which leads to avoiding the learning of the unnecessary motion. 
By the decomposition, four policies are required: Policies A, B, C, and D as shown in \figref{fig:transition}. 
Note that, although a reverse transition also exists, the policies corresponding to this transition can be trained using the proposed training method.

\subsection{Formulation of training}
We formulate the policy training as a Markov Decision Process (MDP). 
Primitive policies corresponding to each transition are learned, and the number of policies is denoted as $N$.
For each transition $i$ $(1 \leq i \leq N)$, the MDP is defined as follows. The MDP has the state $ \bm{s}^{(i)} \in \mathcal{S}^{(i)} $ (where $ \mathcal{S}^{(i)} $ is the state space), action $ \bm{a}^{(i)} \in \mathcal{A}^{(i)} $ (where $ \mathcal{A}^{(i)} $ is the action space), state transition $\mathcal{T}^{(i)}: \mathcal{S}^{(i)} \times \mathcal{A}^{(i)} \rightarrow \mathcal{S}^{(i)} $, initial state distribution $ \rho_0^{(i)} $, and reward function $ r^{(i)}: \mathcal{S}^{(i)} \times \mathcal{A}^{(i)} \rightarrow \mathbb{R} $.
Our objective is to find the policy $ \pi_{\theta}^{(i)} (\bm{a}^{(i)}|\bm{s}^{(i)}) $ that maximizes the cumulative reward $J(\pi_{\theta}^{(i)}) = \mathbb{E}_{\pi_{\theta}^{(i)}} \left[ \sum_{t=0}^{T-1} \gamma^{t} r^{(i)} \left( \bm{s}^{(i)}_{t}, \bm{a}^{(i)}_{t} \right) \right]$ for a given discount factor $\gamma \in [0, 1) $. 
Here, $T$ denotes the episode length, $\bm{s}^{(i)}_{0} \sim \rho_{0}^{(i)} $, $\bm{a}^{(i)}_{t} \sim \pi_{\theta}^{(i)} (\bm{s}_{t})$, and $\bm{s}^{(i)}_{t+1} = \mathcal{T}^{(i)} \left( \bm{s}^{(i)}_{t}, \bm{a}^{(i)}_{t} \right)$.
$\pi_{\theta}^{(i)}$ is parameterized by a weight $\theta$ of a neural network.

The target manipulation is achieved by sequentially executing the policies $\pi^{(i)}$. At this time, the initial state when switching policies is given by $\bm{s}^{(i+1)}_0 = \mathcal{T}^{(i)} \left( \bm{s}^{(i)}_{T-1}, \bm{a}^{(i)}_{T-1} \right)$.

\subsection{Training design}
First, we explain the states and actions required to implement APriCoT. After that, we describe the proposed reward design, which can be commonly reused for all primitives, and initial-state distribution. Subsequently, we explain the overview of the training process. To train the policy efficiently, this study adopts the framework of teacher-student learning. The implementation of teacher-student learning follows the existing research~\cite{qi2023hand}.

\subsubsection{State, Action design}
Toward the application of policies in the real-world, it is desirable to train policies that can adapt to various-shaped objects. For the training, recognizing the shape from observed information is required. In fact, the policy can implicitly utilize the shape information from the difference between the commanded and actual joint angles. Therefore, the state $\bm{s}_t$ is designed to include the commanded joint angles $\bm{q}_t$ and the actual joint angles $\hat{\bm{q}}_t$. The action $\bm{a}_t$ is set as the change in the commanded joint angles $\Delta \bm{q}_t$.

\subsubsection{Reward design}
We design a reward to be reusable for all training of primitive policies. In addition, to promote execution in the appropriate order, we adopt a step-wise reward structure, where the reward incrementally increases as each representation is achieved.
Equation \eqref{eq:reward} represents the reward which is decomposed into terms $r_{\rm transition}$ related to conact-state transitions, and terms $r_{\rm obj}$ related to maintaining the object's position and orientation.
\begin{equation}
    \begin{split}
    r = r_{\rm transition} + r_{\rm obj}
    \end{split}
    \label{eq:reward}
\end{equation}

The term $r_{\rm transition}$ is expressed by equation \eqref{eq:transition}.
\begin{equation}
    \begin{split}
    r_{\rm transition} = w_{\rm det}r_{\rm det} + r_{\rm X} + w_{\rm att}r_{\rm att}
    \end{split}
    \label{eq:transition}
\end{equation}
The components $r_{\rm det}, r_{\rm X}, r_{\rm att}$ constituting $r_{\rm transition}$ depend on the contact-state of finger $F$ moving to the opposite side of the object. $r_{\rm det}$ provides a reward to encourage detach until the finger crosses over the object, where $r_{\rm det}=-d^z_{F}$, and becomes $0$ after the crossover ($d^z_{F}$ is the distance from the object's upper surface to the finger). $r_{\rm X}$ incentivizes crossover motion. This term is a positive constant reward $c_{\rm X1}$ when the finger spans above the object's upper surface; and $c_{\rm X1}+c_{\rm X2}$ when the finger spans below the upper surface; and $0$ when the finger does not cross over the object;  $r_{\rm att}$ encourages attachment. It is $-d^x_{F}$ when the finger spans below the object's upper surface and becomes $-d^x_{F}+c_{\rm att}$ upon achieving the desired contact-state, otherwise $0$ ($d^x_{F}$ is the distance from the object's surface to the finger). Continuous functions for $r_{\rm det}$ and $r_{\rm att}$ increase reward granularity to enhance training efficiency.
$ w_{\rm det}, w_{\rm att}$ are weights of $r_{\rm det}, r_{\rm att}$, respectively.

$r_{\rm obj}$ is represented by equation \eqref{eq:obj}.
\begin{equation}
    \begin{split}
    r_{\rm obj} = w_{\rm dir}r_{\rm dir} + w_{\rm rot}r_{\rm rot} + w_{\rm pos}r_{\rm pos} + r_{\rm term}
    \end{split}
    \label{eq:obj}
\end{equation}
$r_{\rm dir}$ promotes finger-object surface contact for grasp stability. It is $-\sum_{f\neq F} \theta_f$ until the finger spans the object, and \(-\sum\theta_f\) afterward, where $\theta_f$ is the angle between the direction pointed by finger $f$ and the direction perpendicular to the object's upper surface. $r_{\rm rot}, r_{\rm pos}$ prevent changes in the object's position and rotation around axes other than the yaw axis from the initial state. Here, $r_{\rm rot}=-(\theta^{\rm roll}+\theta^{\rm pitch})$, $r_{\rm pos}=-\lVert \bm{p} - \bm{p}^{\rm init}\lVert_2$. $\theta^{\rm roll}, \theta^{\rm pitch}$ are rotation angles around axes penetrating from behind to front and from side to side of the object respectively; $\bm{p}$ is the object's position in the global coordinate system; $\bm{p}^{\rm init}$ is the initial position in the global coordinate system. $r_{\rm term}$ prevents early termination. 
The termination occurs when improving the reward is almost impossible. By the termination, we can enhance the training efficiency. 
It is $c_{\rm term}$ if the episode hasn't terminated, and $-c_{\rm term}$ otherwise. The termination occurs when the object falls from the hand, when finger-surface contact cannot be maintained ($\sum\theta_f > \Theta_1$), or when the object's orientation significantly deviates ($\theta_{\rm roll} > \Theta_2$).
$ w_{\rm dir},  w_{\rm rot},  w_{\rm pos}$ are weights of the reward terms.

\subsubsection{Initial-state design}
\begin{figure*}[t]
  \centering
  \includegraphics[scale=0.3]{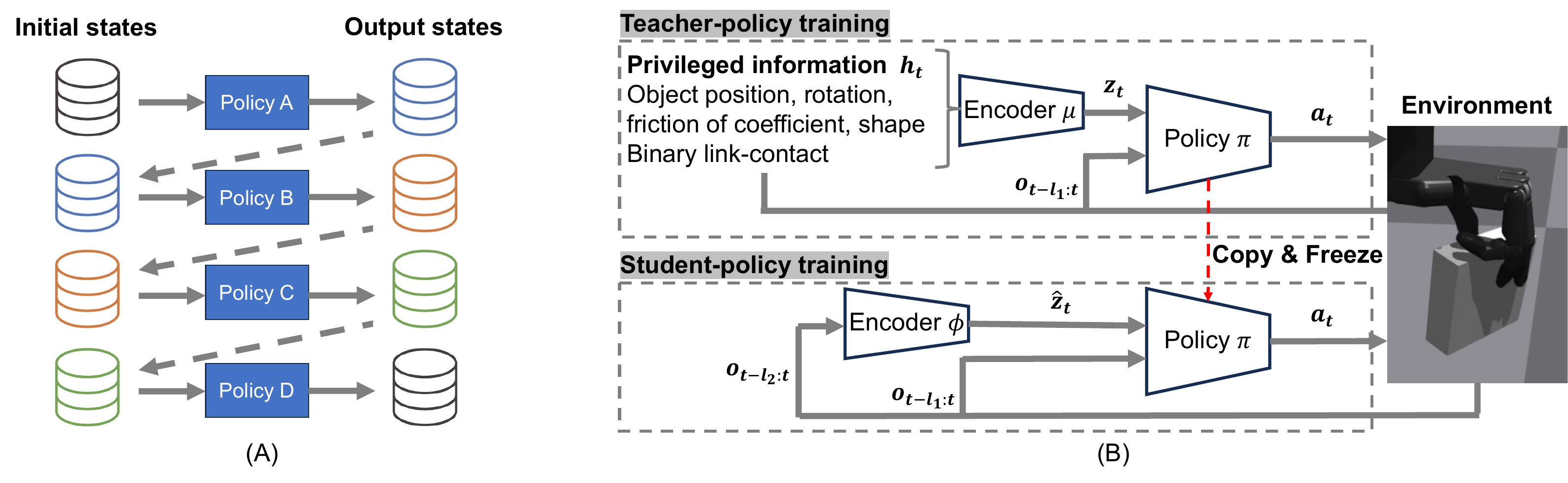}
  \caption{Training overview. (A) explains the initial states used for training. 
  The initial states of Policy B, C, D are the output states of Policy A, B, C, respectively.
  (B) illustrates the steps of teacher-student learning. 
  The teacher policy takes observed information $\bm{o}_{t-l_1:t}$ and a latent variable $\bm{z}_t$, encoded from privileged information $\bm{h}_t$ by the encoder $\mu$, as input to predict an action $\bm{a}_t$. In contrast, the student policy uses $\bm{o}_{t-l_1:t}$ and a latent variable $\hat{\bm{z}}_t$, with $\hat{\bm{z}}_t$ being encoded from $\bm{o}_{t-l_2:t}$ by the encoder $\phi$.}
  \label{fig:training}
  \vspace{-3mm}
\end{figure*}

During the training, diverse initial states are crucial for the robustness of policies. Furthermore, from the perspective of enhancing the overall performance of policies in skill chaining, aligning the input distribution of policies with the output distribution of preceding stages is important \cite{konidaris2009skill, clegg2018learning}. Therefore, we initially train a policy using a large set of initial states generated through simulation. Subsequently, in subsequent policies, we use the states outputted by the preceding policy as initial states (\figref{fig:training} right).

The generation of initial states through simulation follows the following steps:
\begin{enumerate}
    \item Add noise sampled from a uniform distribution to the initial pose of objects, as well as to joint commands.
    \item Advance the simulation for a sufficient number of timesteps ($50$ iterations). During the simulation, we input the same joint commands.
    \item Select initial states where the grasp is maintained as candidates for adoption.
\end{enumerate}
Such collected initial states, generated through these steps, are utilized for the training of Policy A.

\subsubsection{Implementation}
The training utilizes the framework of teacher-student learning as in existing studies~\cite{chen2020learning, lee2020learning, qi2023hand, qi2023general}.
Training a policy from scratch based solely on the observed information is difficult due to the limited object-related information. Training with only the observed information needs a large amount of samples to map the observed with object information, resulting in sample inefficiency.
To mitigate the sample inefficiency, the framework is adopted.
In teacher-student learning, a teacher policy is first trained using the privileged and observed ones. Subsequently, the student policy is trained using only the observed, guided by the teacher policy. This approach facilitates efficient learning.
The teacher and student policy consist of a common policy $\pi$ and different encoders $\mu$ and $\phi$ (Fig. \ref{fig:training}-(B)). 
The encoders $\mu, \phi$ encode the privileged and the multiple observed ones into latent variables $\bm{z}_t, \hat{\bm{z}}_t$, respectively.
During the training of the student policy, the objective is to minimize the following $L2$ loss function $L$, encouraging the encoders $\phi$ to align these latent variables:
\begin{equation}
    L = \|\bm{z}_t - \hat{\bm{z}}_t\|_2^2
\end{equation}
This facilitates learning the correspondence between the privileged and the observed information.

The privileged information $\bm{h}_t$ has object information including the pose, shape, friction coefficients, and binary-contact information between each finger link and the object. To represent the shape, parameters of superquadrics~\cite{barr1981superquadrics} are used. The observed information $\bm{o}_t$ includes commanded joint angles $\bm{q}_t$, measured values $\hat{\bm{q}}_t$, and the action $\bm{a}_{t-1}$ from the previous timestep. 
Action $\bm{a}_t$ includes the change in commanded joint angles $\Delta \bm{q}_t$. 
In the teacher policy, privileged information $\bm{h}_t$ is input to an encoder $\mu$ consisting of fully connected layers, which outputs latent variable $\bm{z}_t$. Subsequently, $\bm{s}_t = \{\bm{z}_t, \bm{o}_{t-l_1:t}\}$ is input to policy $\pi$ consisting of fully connected layers, which outputs action $\bm{a}_t$.
In the student policy, $\bm{o}_{t-l_2:t}$ is input to an encoder $\phi$ constructed with a Temporal Convolutional Network~\cite{lea2017temporal}, which outputs $\hat{\bm{z}}_t$. Then, $\bm{s}_t=\{\hat{\bm{z}}_t, \bm{o}_{t-l_2:t}\}$ is input to $\pi$, which outputs action $\bm{a}_t$.
This study uses $l_1=2, l_2=29$.

Toward the application of the policies in the real world, this study randomizes object shapes, friction coefficients, and measured joint angles. Object shape and friction coefficients are varied when initializing environments. 
The shape parameters are varied using the parameters of superquadrics. Observed joint angles are randomized by adding noise sampled from a uniform distribution at each timestep.

\section{Experiment}
\subsection{Settings}
To rapidly collect a large amount of data, we utilized the IsaacGym Simulator~\cite{makoviychuk2021isaac} for training. During the training process, data collection was achieved by running $32,768$ environments in parallel on NVIDIA RTX3090 GPU. Each environment included a four-fingered hand, specifically the Shadow Dexterous Hand Lite\footnote{https://www.shadowrobot.com/dexterous-hand-series/}, and a various-shaped object. The simulation frequency and control frequency were set at $120$ Hz and $20$ Hz, respectively. Each policy was trained for a maximum of $T=100$ iterations per episode.
The parameter range utilized for the randomization of object shape, friction coefficients, and observed joint angles are shown in \tabref{tab:rand}.
The parameters of the superquadrics include an object depth, width, height, $\epsilon_1$, and $\epsilon_2$. $\epsilon_1$ and $\epsilon_2$ determine the shape, with $\epsilon_2$ altering the shape of top-surface. As $\epsilon_2$ approaches $0$, the shape becomes more rectangular, while it turns more elliptical as $\epsilon_2$ approaches $1$. 
The policy $\pi$ and the encoder $\mu$ multi-layer perceptrons (MLP) with hidden layers consisting of $[512, 256, 128, 64]$ and $[256, 64]$ units, respectively. We use the exponential linear unit (ELU)~\cite{clevert2015fast} as the activation function for these networks. The structure of encoder $\phi$ is the same as used in~\cite{qi2023hand}.
The Proximal Policy Optimization (PPO) algorithm~\cite{schulman2017proximal} was used for training the teacher policy. The student policy was optimized using the Adam optimizer~\cite{kingma2014adam}.
The hyperparameters for reward and early termination and reward coefficients used for each policy were set as shown in \tabref{tab:table1}.

\begin{table}[t]
\caption{The range of parameter randomization in the experiment}
\label{tab:rand}
\makebox[0.45 \textwidth][c]{
\begin{tabular}{cc}
Parameter             & Range                         \\ \hline
Object depth          & $0.02, 0.0275, 0.35$ [m]    \\
Object width          & $0.11$ [m]                         \\
Object height         & $0.1$ [m]                          \\
Object shape ($\epsilon_1$)     & $\simeq 0$                         \\
Object shape ($\epsilon_2$)     & $\simeq 0, 0.5, 1$       \\
Friction coefficients & uniform([$0.7, 1.3$])       \\
Joint noise           & uniform([$-0.05, 0.05$]) [rad]
\end{tabular}
}
\end{table}

\begin{table}[t]
\caption{Hyperparameters of reward and early termination.}
\label{tab:table1}
\makebox[0.5 \textwidth][c]{
\begin{tabular}{c|cccccc}
            & $c_{\rm X1}$ & $c_{\rm X2}$ & $c_{\rm att}$ & $c_{\rm term}$ & $\Theta_1$                     & $\Theta_2$ \\ \hline
A,B,C & 0.1 & 0.2 & 1 & 0.1 & $120^{\circ}$ & $15^{\circ}$\ \\
D  & 0.2 & 0.5 & 0.5 & 0.1 & $120^{\circ}$ & $15^{\circ}$
\end{tabular}
}

\makebox[0.5 \textwidth][c]{
\begin{tabular}{c|cccccc}
            & $w_{\rm det}$ & $w_{\rm att}$ & $w_{\rm dir}$ & $w_{\rm rot}$ & $w_{\rm pos}$                  \\ \hline
A,B,C & 0.1 & 0.1 & 0.1 & 0.1 & 0.01\ \\
D  & 2 & 1 & 0.1 & 0.1 & 0.01
\end{tabular}
}
\end{table}


To demonstrate the effectiveness of the proposed method for the target manipulation, we compared it against the following two baseline methods in the experiments:
\begin{enumerate}
\item \textbf{Baseline A}: A modified version of the method that targets only the rotation of objects within the hand~\cite{andrychowicz2020learning}. A positive reward given when realizing the grasp was simply added to the original reward. By comparing this baseline with ours, we verify that the intended grasp can be achieved more effectively by considering the transitions and decomposing into the primitives.
\item \textbf{Baseline B}: A method designed for pinchable objects~\cite{qi2023hand}. This baseline was compared with ours to investigate that learning the manipulation requires a wide exploration, and that the decomposition into short-term primitives makes learning feasible. In this method, a negative reward is added for changes in joint angles, encouraging the rotation of the object without significantly altering the joints. As a result, this method can rotate objects while maintaining the desired grasp only for pinchable objects.
\end{enumerate}

\subsection{Results}
\subsubsection{Comparison}
\begin{figure*}[t]
  \centering
  \includegraphics[scale=0.5]{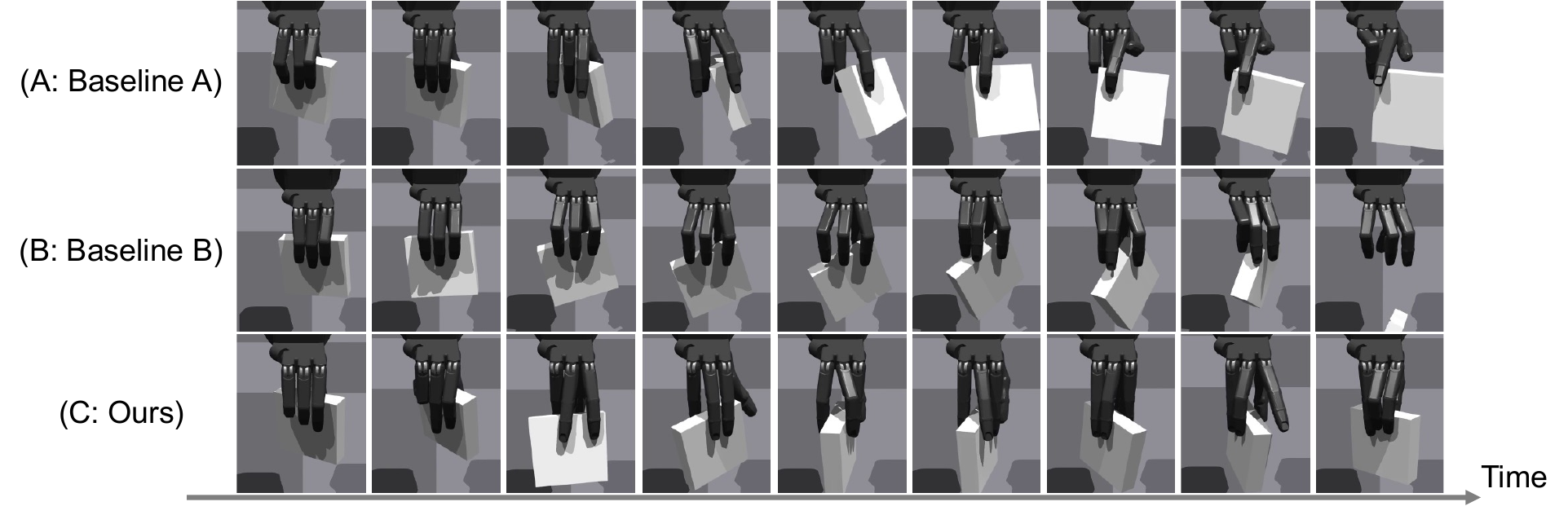}
  \caption{The example of results. (A):Baseline A, (B):Baseline B, (C):Ours. At the final timestep, (A) achieved to rotate the object but the intended grasp is not realized. In (B), the object is fallen. On the other hand, (C) achieved the rotation and desired grasp.}
  \label{fig:comparison}
  \vspace{-5mm}
\end{figure*}

\figref{fig:comparison} shows the training results of the baselines and the proposed method. 
The success is defined as achieving the target grasp after rotating the object by half a turn at the final timestep.
The baselines failed to train a policy achieving both the in-hand half-rotation and the desired grasp.
In Baseline A, the rotation was achieved, but the intended grasp was not realized. This is because a long step results in sparse rewards on the successful grasp.
In Baseline B, it rotated the object around the roll direction for crossovers, reducing the contact area and causing the object to drop due to the inability to maintain the grasp. 
This is caused by limiting the exploration of joint angles through the reward. This limitation makes it difficult to perform crossover while preventing the object rotation around the roll direction.
In contrast, our method successfully achieved the desired grasp after rotating the object.
We believe that designing the primitives based on the contact-state transition helped achieve the target grasp and the decomposition simplified the exploration of the actions to perform crossover.

\subsubsection{Robustness}
\begin{figure}[t]
  \centering
  \includegraphics[width=\columnwidth]{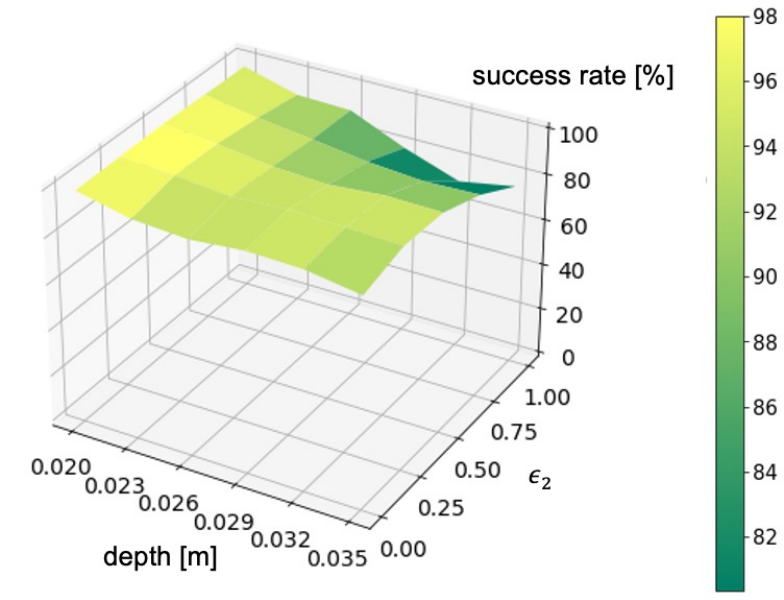}
  \caption{Success rate. The vertical axis represents the success rate for each object shape. The axes of the base plane represent the depth and $\epsilon_2$ of the used objects.}
  \label{fig:success_rate}
  \vspace{-3mm}
\end{figure}

To evaluate the robustness of the proposed policies, this study assessed the success rates of operations for different object shapes. 
The depth and $\epsilon_2$ are selected from $\{0.02, 0.023, 0.026, 0.029, 0.032, 0.035\}[m]$ and $\{\simeq 0, 0.25, 0.5, 0.75, 1\}$, respectively.
We conducted the evaluation by randomly sampling $1,000$ initial states and calculating the success rates for each shape. The success rates for each shape are presented in \figref{fig:success_rate}. We achieved success rates of over $90\%$ for almost all shapes. The results indicate that our policies are robust to changes in object shape.
The success rate with the unused shapes in the training was almost the same as with the used shapes.
For objects with a large depth and a high $\epsilon_2$ (thicker ellipsoids), the success rate was around $75\%$. This lower success rate is attributed to the increased curvature, making the objects more prone to slipping sideways during manipulation.

\subsubsection{Visualization of latent variables}
\begin{figure}[t]
  \centering
  \includegraphics[width=\columnwidth]{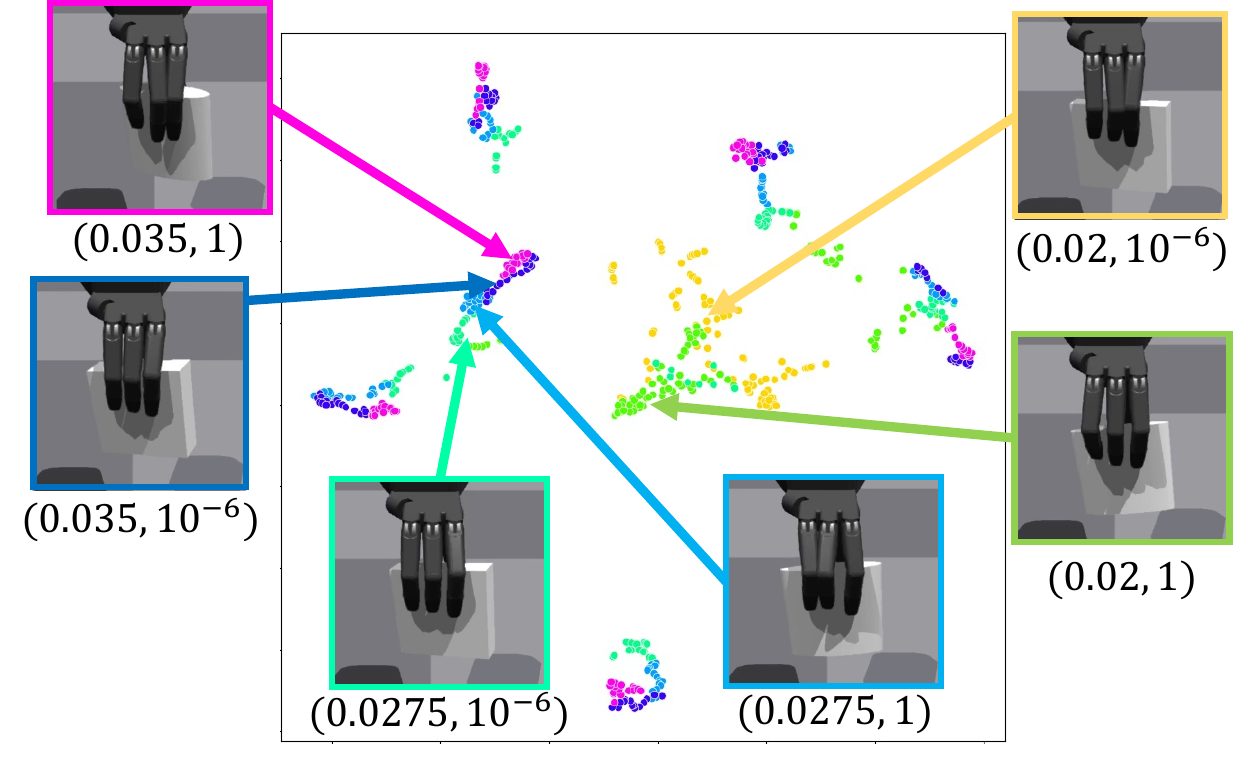}
  \caption{Visualization of latent variables from encoder $\phi$ using t-SNE. The points represent variables collected during policy execution with six objects. The object depth and $\epsilon_2$ are written below each image. Distinct colors correspond to different objects. As seen in arrowed areas, points with the same color are clustered closely together.}
  \label{fig:latent}
  \vspace{-5mm}
\end{figure}

To verify whether the features of different object shapes were learned, we visualized the distribution of latent variables $\bm{\hat{z}_t}$ using t-SNE~\cite{van2008visualizing} (\figref{fig:latent}). 
The depth and $\epsilon_2$ are chosen from $\{0.02, 0.0275, 0.035\}[m]$ and $\{\simeq 0, 0.5, 1\}$, respectively.
The latent variables were collected by running Policy A for $100$ iterations. It was observed that points corresponding to similar shapes were clustered closely together. 
This suggests that the features of the object shapes were learned by the policy, resulting in the policy being robust to changes in the shapes.

\section{Discussion}
\subsection{Summary of the experiments}
This study proposed APriCoT to execute in-hand tool manipulation. Experiments showed that while the baselines could not achieve both in-hand rotation and the desired grasp, our method successfully achieved the target grasp after the rotation.
Even with the known contact-state after manipulation, existing methods struggle with long-term operations.
Our approach simplifies learning by decomposing manipulation based on the transition. 
By this decomposition, we can train the policy to learn each primitive where fingers need to just perform short-term and similar motions, which improves sample efficiency.
In addition, the experiment demonstrated that the policies were robust to the changes in the object shape. This robustness is important to apply the policies in the real-world where various-shaped objects exist.

Visualization of latent variables suggests the policy implicitly recognizes the object shape from joint angles and adapts actions accordingly. However, information of object pose, such as roll rotation, cannot be detected solely from joints. Failure to detect this rotation can destabilize and drop the object.
In addition, for tasks with a goal pose of the object, precisely achieving the goal after the manipulation is challenging. Focusing solely on contact-state transitions, as in this study, does not ensure the desired orientation. Therefore, policies must adjust actions based on observations recognizing object orientation.
In the future, we integrate the existing methods using visual information~\cite{andrychowicz2020learning, handa2023dextreme, qi2023general} with ours to deal with this issue.


Although we conducted only simulation experiments, we believe that the proposed policies are applicable in the real-world. Following previous research that demonstrated feasibility in the real, we employed domain randomization and the framework of teacher-student learning for adaptation during policy training. These techniques might help bridge the sim-to-real gap. 

\subsection{Realizing diverse manipulation using reusable primitives}
This study designed action primitives for rotating an elongated object to achieve a precision grasp. 
Various grasp-types exist other than a precision grasp, and achieving such diverse grasps by in-hand tool manipulation requires many policies.
An example of these operations is achieving a power grasp by moving the object toward the palm after the manipulation which was dealt with in this study.
Common contact-state transitions occur in various manipulation. Policies for specific transitions should be reusable across operations. By focusing on reusability 
a wide range of operations could be achieved, reducing the number of required primitives.
This study is a first step toward enabling such various operations by leveraging action-primitive reusability.

This study executed primitives based on the contact-state transition-graph. Manually designing the graph for all operations is labor-intensive. This could be mitigated by automating graph construction via methods like Learning-from-Observation, which uses human demonstrations for task-sequence construction \cite{ikeuchi1994toward, wake2021learning, saito2022task, wake2023text}. 
Additionally, we can deal with the issue using a learning-based method, such as hierarchical reinforcement learning~\cite{pateria2021hierarchical} to train a high-level policy for choosing primitives based on the current state.

\section{Conclusion}
To achieve the skill of in-hand tool manipulation that not only manipulates a tool within the hand but also achieves a grasp suitable for a task after the manipulation, we proposed Action Primitives based on Contact-state Transition (APriCoT). 
APriCoT decomposes the operation into several action primitives. In each primitive, fingers need to just perform short-term and similar motions, which simplifies the training of primitive policies using DRL. 
Experiments demonstrated that our method achieves both in-hand rotation and the desired grasp using various-shaped objects.
Our method has potential for other operations, provided primitive policies for different contact-state transitions are prepared. This study is a first step toward achieving diverse operations.

\section*{ACKNOWLEDGMENT}
This work was supported by Grant-in-Aid for JSPS Fellows Grant Number JP24KJ1054.

\bibliographystyle{unsrt}
\bibliography{bib}

\end{document}